\documentclass{article}

\usepackage[preprint]{neurips_2024}

\usepackage{csquotes}
\usepackage{paralist}
\usepackage{adjustbox}
\usepackage{enumitem}
\usepackage{graphicx}
\usepackage{booktabs}
\usepackage[utf8]{inputenc} 
\usepackage[T1]{fontenc}    
\usepackage{hyperref}       
\usepackage{url}            
\usepackage{booktabs}       
\usepackage{amsfonts}       
\usepackage{nicefrac}       
\usepackage{microtype}      
\usepackage[dvipsnames]{xcolor}         
\usepackage{tikz}
\usepackage{doi}

\usetikzlibrary{positioning, arrows.meta,decorations.pathreplacing,calc,shapes.geometric}

\setcitestyle{numbers,square}
\title{LLM2Swarm: Robot Swarms that Responsively Reason, Plan, and Collaborate through LLMs}

\author{%
  Volker Strobel\\
  IRIDIA\\
  Universit\'e Libre de Bruxellles\\
  Brussels, Belgium \\
  \texttt{volker.strobel@ulb.be} \\
  \And
  Marco Dorigo\\
  IRIDIA\\
  Universit\'e Libre de Bruxellles\\
  Brussels, Belgium \\
  \texttt{mdorigo@ulb.ac.be} \\
  \AND
  Mario Fritz\\
  CISPA Helmholtz Center for Information Security\\
  St. Ingbert, Germany\\
  \texttt{fritz@cispa.de}
}

\newif\ifcmnts
\cmntstrue

\newcommand{\myparagraph}[1]{\paragraph{#1}}

\renewcommand{\myparagraph}[1]{}

\newif\ifcomments
\commentstrue

\newif\ifshowresolved
\showresolvedfalse

\newcommand{\rmario}[1]{\ifshowresolved{\color{blue}{mario: #1}}\fi}
\newcommand{\rvolker}[1]{\ifshowresolved{\color{Green}{volker: #1}}\fi}

\begin{document}
\maketitle

\begin{abstract}
\rmario{concerning title "LLM-enhanced" sounds very incremental - as if we won't expect any qualitative improvements or disruptive changes. maybe "LLM-enabled" is better? I'm trying different rephrasings I'd also brand this as "LLM2Swarm"} \rvolker{LLM2Swarm sounds good to me. Also LLM-enabled is fine for me  -- it first sounded a bit strange but then I realized that it is very similar to other uses, like in `Bluetooth-enabled device'. An alternative to LLM-enabled could be LLM-integrated as `integration' is the word we use most-often in the paper.}

Robot swarms are composed of many simple robots that communicate and collaborate to fulfill complex tasks. Robot controllers usually need to be specified by experts on a case-by-case basis via programming code. This process is time-consuming, prone to errors, and unable to take into account all situations that may be encountered during deployment. On the other hand, recent Large Language Models (LLMs) have demonstrated reasoning and planning capabilities, introduced new ways to interact with and program machines, and incorporate both domain-specific and commonsense knowledge. Hence, we propose to address the aforementioned challenges by integrating LLMs with robot swarms and show the potential in proofs of concept (showcases). For this integration, we explore two approaches. The first approach is `indirect integration,' where LLMs are used to synthesize and validate the robot controllers. This approach may reduce development time and human error before deployment. Moreover, during deployment, it could be used for on-the-fly creation of new robot behaviors. The second approach is `direct integration,' where each robot locally executes a separate LLM instance during deployment for robot-robot collaboration and human-swarm interaction. These local LLM instances enable each robot to reason, plan, and collaborate using natural language, as demonstrated in our showcases where the robots are able to detect a variety of anomalies, without prior information about the nature of these anomalies. To enable further research on our mainly conceptual contribution, we release the software and videos for our LLM2Swarm system: \url{https://github.com/Pold87/LLM2Swarm}.
\rmario{I'm still not sure about "indirect" vs "direct". could it also be "pre-mission" and "in-mission"? I'll read on and come back to this.} \rvolker{I am not sure either and am open for suggestions. Previously, I had `offline' and `online', with the meaning of `pre-mission' and `in-mission'. However, I think the real difference lies in whether or not you use the LLM output directly or via the indirect route of controller generation/synthesis.}\rmario{yes - maybe then it's more "synthesis-based" vs "interactive"; well - maybe "direct" vs "indirect" integration is not so bad actually}
\end{abstract}

\section{Introduction}

\rmario{high level comment: i like the reason, plan, collaborate slogan .... but it should also reappear somewhere in the submission. E.g. text or structuring elements - like headings}\rvolker{I added the slogan to Opportunities.}

Robot swarms consist of many simple robots that accomplish complex tasks by collaborating with each other~\cite{DorTheTri2020:scirobotics,Ham2018:book}. They are characterized by the absence of a central control unit, which makes them more scalable and flexible than other multi-robot systems. Thanks to their decentralized control and redundancy, robot swarms can potentially continue functioning even if individual robots fail~\cite{WinNem2006:mic}. These features could make robot swarms ideal for applications where fast reactions to unforeseen events are required, such as environmental monitoring and disaster response~\cite{DorTheTri2021:pieee,YanBelDup-etal2018:scirobotics}.

\rmario{the statement "often simple robots" is a key requirement/point in your argumentation - and I'm afraid it might be overlooked. you need to make it more prominent and not optional (as in "often") - I think. You need it in multiple places below: (1) clearly delimit your work from prior work on general robots with LLMs (2) motivate the control synthesis for responsiveness.} \rvolker{I was a bit afraid of using 'simple' because people in swarm robotics then might think of Kilobots, which have a 8MHz processor. and will never be able to run LLMs... but I agree, it is one of the main selling points and I added 'simple' to the first sentence and removed this sentence as it was redundant.}

In robot swarms, communication and interaction protocols between robots usually need to be specified by experts on a case-by-case basis via programming code~\cite{Kuc2023:frontiers}: Given a specific task, a human designer needs to develop and implement the corresponding program logic. The process of writing such code is often time-consuming and prone to errors. Even though there are attempts to automate this process~\cite{FraBraBru-etal2014:si,GarBir2024:icra,TriTucAmpDor2014:evorobotics,TolGamPaul-etal2020:roblearning,FinMetMos2022:cav}, these attempts usually still require expert knowledge at design time  and the controllers cannot react to unforeseen events at deployment time.

Recent advances in Large Language Models (LLMs) have led to new possibilities and proposals for the integration of AI into different technologies, such as virtual assistants and search engines~\cite{GreAbdMis-etal2023:aisec}. In particular, there have also been proposals to integrate LLMs into multi-robot systems~\cite{CheArkZha-etal2024:icra,ManJaiSon2023:icra}. With this paper, we propose to integrate LLMs into a specific kind of multi-robot system: robot swarms.

We provide the first systematic exploration of the potential of LLM-enabled robot swarms (we call our system LLM2Swarm). We demonstrate the resulting opportunities in a series of proofs of concept (showcases), implemented in the ARGoS robot swarm simulator~\cite{PinTriOGr-etal2012:si}, and preliminary hardware tests using a Raspberry Pi~5. For this exploration, we propose two primary approaches of integrating LLMs into robot swarms. 

The first approach is ``indirect integration,'' where LLMs are used to synthesize and validate the robot controllers (before and/or during deployment). This approach may save development time for the designer by automating parts of the design process. In addition, it could dramatically simplify the process of writing controllers, as no programming knowledge is necessary. The approach ensures that robot swarms \emph{responsively} react, as they execute their classical (synthesized) control software, instead of waiting for LLM-generated responses.

The second approach for integrating LLMs into robot swarms is ``direct integration,'' where each robot locally executes a separate LLM instance during deployment. With this online integration, the local LLM instances enable each robot to reason, plan, and collaborate using natural language. With these new and improved capabilities, we aim to enhance robot swarms' adaptivity, efficiency, and intelligence, making them more capable of handling complex tasks in unpredictable environments, both for robot-robot collaboration and human-swarm interaction.


In summary, our contributions are as follows:%
\begin{enumerate}
    \item We introduce LLM2Swarm for integrating LLMs into robot swarms and explore the resulting opportunities and challenges.
    \item We provide (video) showcases for controller synthesis, robot-robot collaboration, and human-swarm interaction in simulation.
    \item We release our LLM2Swarm setup as open-source software. Videos and code can be downloaded at \url{https://github.com/Pold87/LLM2Swarm}.
\end{enumerate}

\begin{figure}
    \centering
    \includegraphics[width=0.95\textwidth]{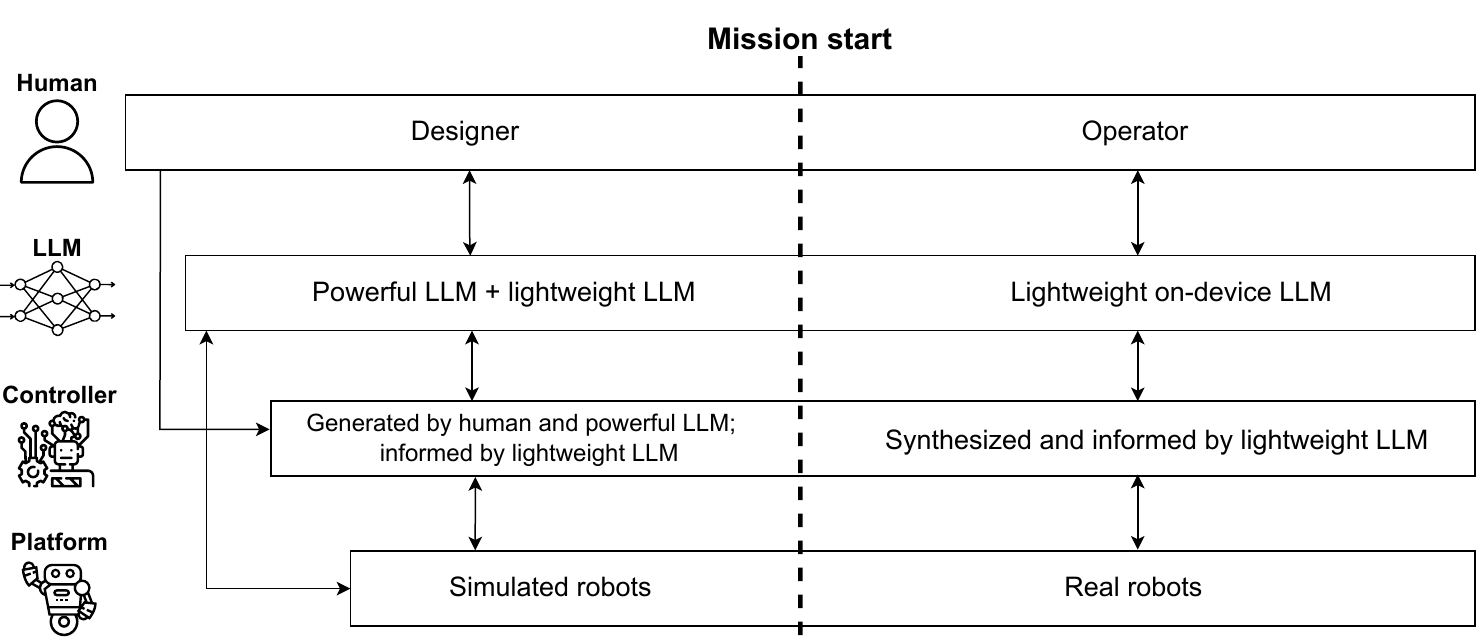}
    \caption{\textbf{Overview of LLM2Swarm -- LLM-enabled robot swarms}. LLM2Swarm involves four key system components: humans, LLMs, controllers, and platforms.
    \emph{Before mission start}: a human designer uses both manual design and LLM2Swarm's controller synthesis module (which prompts a powerful LLM) to generate a robot controller. This controller is executed in simulation, and uses one lightweight LLM per robot to simulate on-device execution of LLM2Swarm's direct integration module. \emph{After mission start}: a human operator interacts with the real robots' lightweight on-device LLMs to instruct the swarm and to receive information about the swarm state. These LLMs also interact with other robots' controllers in order to reason, plan, and collaborate. In addition, the lightweight LLMs can synthesize new robot controllers on-the-fly during the mission.}
    \label{fig:temporal-flow-interactions}
\end{figure}

\section{Related work}
\label{sec:related-work}

Our work is related to three topics: controller synthesis, LLMs as agents, and LLMs in robotics. In the following, we briefly present related work from these fields and highlight how our vision differs.

\textbf{Controller synthesis.}
There are several related works on controller synthesis, both from the field of swarm robotics (e.g., AutoMoDe~\cite{FraBraBru-etal2014:si,GarBir2024:icra}, evolutionary approaches~\cite{TriTucAmpDor2014:evorobotics}, neural network controllers~\cite{TolGamPaul-etal2020:roblearning}) and from more theoretical fields, such as computer aided verification~\cite{FinMetMos2022:cav}. LLMs have also already been used for the controller synthesis of single robots~\cite{VemBonBucKap2024:techrep}. However, these approaches still rely on expert knowledge (e.g., formal specification languages) during design time, cannot flexibly react to unforeseen situations during deployment, or do not support multiple agents or robots.

\textbf{LLMs in multi-agent systems.}
LLMs are increasingly being integrated into other systems to enhance their capabilities (e.g., as part of search engines)~\cite{GreAbdMis-etal2023:aisec}.
In the recent past, there has been an increasing interest in letting \emph{multiple} LLMs interact and discuss~\cite{AbdGomSiv-etal2024:arxiv,ChuGoyHar-etal2023:arxiv}.
An overview of the integration of LLMs into multi-agent systems for various tasks has been given by Guo et al.~\cite{GuoCheWan-etal2024:arxiv}. In contrast to these works, our target platforms are physical robots. The previous works did not take into account how to synthesize robot controllers, how to deal with mobility, sensor information, and actuators, and how to map sensors and actuators to LLM inputs and responses.

\textbf{LLMs in robotics.}
Integrating LLMs into other systems has also led to several new applications in robotics~\cite{KimKimCho-etal2024:intellservicerobotics}, for example for task planning~\cite{DinZhaAmi-etal2023:autorob}, centralized motion planning~\cite{JiaPatKhu-etal2023:neurips}, and human-robot interaction~\cite{VemBonBucKap2024:techrep}.
As our target platforms are robot swarms, in the following, we are considering works that use a separate LLM instance for each embodied agent. 
%
\citeauthor{CheArkZha-etal2024:icra}~\cite{CheArkZha-etal2024:icra} perform motion planning (e.g., to move a box to a target), with up to 32~simulated robots (in 2D grid environments and using robotic arms).
\citeauthor{ManJaiSon2023:icra}~\cite{ManJaiSon2023:icra} perform motion planning (of robotic arms) with up to three robots.
\citeauthor{ZhaDuSha-etal2024:iclr}~\cite{ZhaDuSha-etal2024:iclr} study how two virtual humans can communicate via LLMs. 
These research efforts, however, have one or more of the following limitations: a low number of robots/agents, lack of decentralization, no consideration of mobile robots with physics-based dynamics, no consideration of typical swarm robotics tasks, or no consideration of hardware limitations; therefore, their applicability to swarm robotics is limited.

\section{LLM-enabled robot swarms: LLM2Swarm}

\rmario{there is a section missing right before "Opportunities" that explains our approach to integrating LLMs in swarms. this needs to settle terminology and use figure 1 to describe the overall landscape. I moved a bit of text around (from caption to here and start of opportunities) ... but there needs to be a bit more describing our main idea and the topologie/structure. this is one of our core contributions! this should not be entangled with opportunities and challenges. before - the main method/approach/structure was described in the captions - this is not good.}

To give an overview of how our LLM2Swarm system yields LLM-enabled robot swarms, we illustrate the proposed temporal flow and interactions in \autoref{fig:temporal-flow-interactions}.
LLM2Swarm provides two main modules: an \emph{indirect integration module} for automating controller design (both before and during deployment) and a \emph{direct integration module} for enhancing robots' reasoning, planning, and collaboration capabilities during deployment.

\rvolker{TODO: Check if this can be included somewhere: : the mission start marks a crucial event in the life cycle of LLM-enabled robot swarms: at this point and after, the LLMs that reason, plan, and collaborate need to be executed on-board of real robots.}



\subsection{Indirect Integration}

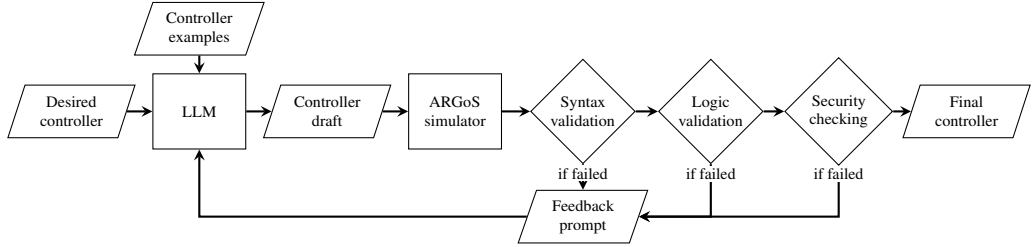
\begin{figure}
    \centering
\begin{tikzpicture}[node distance=1.7cm,font=\tiny]
    \tikzstyle{startstop} = [rectangle, rounded corners, 
minimum width=1cm, 
minimum height=1cm,
text centered, 
text width=1cm,
draw=black]

\tikzstyle{io} = [trapezium, 
trapezium stretches=true, 
trapezium left angle=70, 
trapezium right angle=110, 
minimum width=1.0cm, 
minimum height=0.7cm, text centered, 
text width=1.0cm,
draw=black]

\tikzstyle{process} = [rectangle, 
minimum width=1cm, 
minimum height=1cm, 
text centered, 
text width=1cm, 
draw=black]

\tikzstyle{decision} = [diamond, 
minimum width=0.7cm, 
minimum height=1cm, 
text centered, 
text width=1.0cm,
inner sep=0pt,
draw=black]
\tikzstyle{arrow} = [thick,->,>=stealth]

\node (in1) [io] {Desired controller};
\node (llm) [process, right of=in1, xshift=0cm] {LLM};
\node (examples) [io, above of=llm, yshift=-0.6cm] {Controller examples};
\node (controllerdraft) [io, right of=llm] {Controller draft};
\node (argos) [process, right of=controllerdraft] {ARGoS\\simulator};
\node (syntaxvalidation) [decision, right of=argos] {Syntax validation};
\node (logicvalidation) [decision, right of=syntaxvalidation] {Logic validation};
\node (securitychecking) [decision, right of=logicvalidation] {Security checking};
\node (finalcontroller) [io, right of=securitychecking] {Final controller};
\node (feedback) [io, below of=syntaxvalidation,yshift=0.3cm] {Feedback prompt};



\draw [arrow] (in1) -- (llm);
\draw [arrow] (examples) -- (llm);
\draw [arrow] (llm) -- (controllerdraft);
\draw [arrow] (controllerdraft) -- (argos);
\draw [arrow] (argos) -- (syntaxvalidation);
\draw [arrow] (syntaxvalidation) -- (logicvalidation);
\draw [arrow] (logicvalidation) -- (securitychecking);
\draw [arrow] (securitychecking) -- (finalcontroller);
\draw [arrow] (syntaxvalidation) -- node[yshift=0.05cm,fill=white,inner sep=0.8pt] {if failed} (feedback);
\draw [arrow] (logicvalidation) |- node[yshift=0.57cm,fill=white,inner sep=0.8pt] {if failed}(feedback);
\draw [arrow] (securitychecking) |- node[yshift=0.57cm,fill=white,inner sep=0.8pt] {if failed} (feedback);
\draw [arrow] (feedback) -| (llm);

\end{tikzpicture}
    \caption{\textbf{Flow of LLM2Swarm's controller synthesis module}. Using LLM2Swarm's controller synthesis module, a user begins by specifying the desired controller in natural language. This specification, together with controller examples, is used as part of an LLM prompt to synthesize a robot controller. The synthesized controller draft is then executed directly in the ARGoS simulator. If ARGoS detects any syntax errors, the errors are reported back to the LLM, with the request to resolve them. If no syntax errors are found, the user can proceed to validate the controller logic: if the robot behavior is not as expected, the user provides information about both the robots' expected behavior and actual behavior to the LLM with the request to improve the controller. Once the controller logic is validated, the user can ask the LLM to check for any security vulnerabilities. After all validation steps are completed, the final controller, specified in programming code, is ready for deployment.}
    \label{fig:controller-generation}
\end{figure}

We propose to use LLMs `indirectly' for controller synthesis, i.e., automated creation of code before or during deployment. For this indirect integration, LLM2Swarm provides a module (see \autoref{fig:controller-generation}) that translates a description of a robot controller in natural language to an actual robot controller.
Before the mission start, a human designer generates a robot controller draft using manual design. Parts of these controllers can be synthesized by LLM2Swarm, using a powerful LLM (LLM2Swarm supports any model provided by OpenAI's API or by Ollama for local execution).
Also during deployment, this LLM2Swarm module can lead to increased autonomy of a robot swarm---as missing controller parts can be synthesized on the fly by on-board LLMs.
Based on our experiments, we propose the following three-step process to automate controller synthesis:%
\paragraph{Step~1 -- Syntax validation.} The first validation is to check whether a synthesized controller contains syntactically correct and executable code.
Consequently, a syntax validator is essential to prevent the execution of erroneous code, which could lead to disastrous outcomes (e.g., erroneous code controlling a drone could result in crashes and severe hardware damage).

\paragraph{Step~2 -- Logic validation.}Validating whether the generated code performs the intended task as expected is crucial but challenging, as there is not always a clear ground truth. If the code is synthesized before deployment, its resulting behavior can be analyzed and improved using LLM2Swarm---either by human feedback or in the future by the video analysis capabilities of LLMs 
(although the current video analysis capabilities are limited\footnote{Even though OpenAI's API does not yet support video analysis, in a preliminary experiment, 
we tested ChatGPT-4o's video analysis capabilities by uploading a video of \emph{random-walk} behavior in an ARGoS simulation and asking whether the swarm demonstrated successful \emph{flocking} behavior. The LLM's analysis was unsatisfactory, as it incorrectly concluded that the behavior was successful flocking behavior.}, we believe that future developments will improve this capability). During deployment, however, such verification is usually not possible---in addition, simulations cannot account for all possible real-world scenarios. Alternatives for evaluating a controller during deployment could be reinforcement learning to compare different controllers~\cite{BlaAkh2023:cognitiverobotics} or probabilistic model checking based on mathematical analysis~\cite{KonDixFis2021:ras}.

\paragraph{Step~3 -- Security checking.} It is important to verify whether the LLM-synthesized controller contains security issues or potentially malicious parts (e.g., the controller could contain logic to sabotage other robots). Identifying security issues can be based on the \emph{Common Weakness Enumeration} (CWE), a catalog that lists known exploits in programming languages and hardware components\footnote{\url{https://cwe.mitre.org/}, accessed September 20, 2024}. In addition, there are model checkers and LLMs fine-tuned to the task of identifying weaknesses that are part of the CWE. LLM2Swarm enables preliminary security checking through an LLM prompt with the request of identifying malicious parts and other weaknesses in a synthesized controller.

\subsection{Direct Integration}

\begin{figure}
    \centering
    \includegraphics[width=0.85\linewidth]{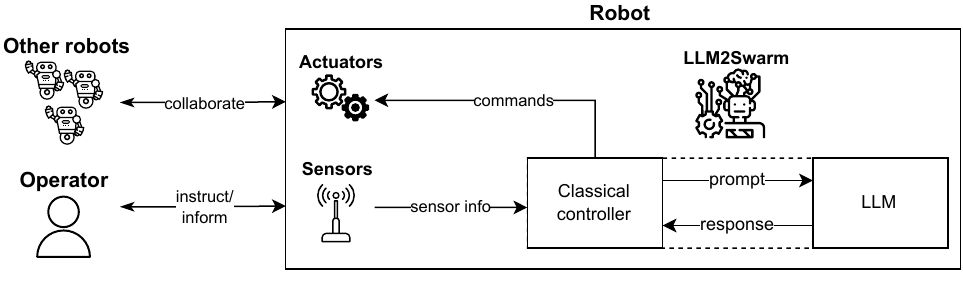}
    \caption{\textbf{System interactions for LLM2Swarm's direct integration module.} Using LLM2Swarm's direct integration module, a robot's controller is composed of two parts: a classical controller and an on-device LLM. As in traditional approaches, the classical controller manages the robot's actuators and sensors. However, unlike traditional approaches, the controller also creates prompts for the on-device LLM and uses the responses to guide the robot's actions. This on-device LLM also enables a robot to interact with other robots or a human operator by using natural language. As a result, such LLM-enabled robots can reason, plan, and collaborate thanks to the capabilities of their on-device LLMs.}
    \label{fig:system-interaction}
\end{figure}

For directly integrating LLMs into robot swarms, we propose to execute one LLM per robot, locally on each robot's hardware (see \autoref{fig:system-interaction}), which requires lightweight LLMs. Currently, LLM2Swarm \emph{simulates} local execution: In our showcases, each robot is connected to a separate \emph{external} LLM, allowing us to run LLMs of any capacity and thus simulate today what we anticipate for the future.
For direct integration, the LLMs' prompts are fed with sensor information (including messages received from other robots), and the outputs of the LLMs are mapped to robot actions. 
During deployment, a human operator may interact with these on-board LLMs to supervise the mission.

\subsection{Experiment setup}
\label{sec:experiment-setup}


In our showcases, we simulate Pi-puck robots using the widely-used robot swarm simulator ARGoS~\cite{PinTriOGr-etal2012:si} (version 3.0.0-beta56) on a personal computer (OS: Ubuntu 22.04.4 LTS, CPU: Intel Core i7-8550U\,@\,1.80~GHz, RAM: 16~GB, GPU: Intel UHD Graphics 620). Although ARGoS controllers are usually written in C\texttt{++}, we use the ARGoS-Python interface~\cite{HasParPacStr2024:software} (version v1.0.0), enabling us to easily connect each simulated robot to an LLM instance (we use GPT-4o). 


\section{Opportunities}
\label{sec:opportunities}

\rmario{short outline of the opportunities section}

In this section, we explore the opportunities that are enabled by LLM2Swarm: automated controller synthesis, enhanced robot-robot collaboration, and more intuitive human-swarm interaction.

\subsection{Automated controller synthesis}
\label{sec:automated-controller-generation}

Synthesizing controllers maintains the \emph{responsiveness} of a robot swarm, as this can be done before deployment or during a low-activity phase during deployment. Overall, this automated controller synthesis can lead to faster development times, the specification of controllers by non-experts, and on-the-fly adaptation of robot controllers during deployment.

For creating controllers by LLMs, different design choices need to be made. It needs to be decided whether the LLM should synthesize the complete controller or only specific parts (e.g., only the movement logic). Another design choice is whether the controller (parts) should be written by an LLM before or rather during deployment. These design choices balance the trade-off between autonomy and controllability (see also \autoref{sec:controllability}).


\subsection{Robot-robot collaboration and human-swarm interaction}
\rmario{make sure that you always describe it in the swarm context}
\rmario{this subsection was a huge chunk of text. this is not easy for the reader. I've tried to give some more structure with the "paragraph" tags. in general, paragraphs should be shorter.}


\paragraph{Robot-robot collaboration and adaptivity in swarms.}

We believe that LLM2Swarm 
can largely improve robot-robot collaboration. First, enhancing robots with LLMs expands their ability to \emph{reason}. 
Pre-trained LLMs incorporate general knowledge about the world and context-aware reasoning, which they can use to react to unanticipated challenges, incorporating moral values and favoring ethical actions---potentially considering how humans would react in similar situations.

Second, LLM2Swarm provides robots with advanced \emph{planning} capabilities. Pre-trained LLMs can generate collective plans, involving the allocation of different robots to different tasks and the ability to adapt plans in real-time (for example, to adapt a path planning procedure when unanticipated situations occur). LLMs are also able to analyze the history of interactions and decide on the best next collective action based on this.

Finally, LLM2Swarm enables robots to \emph{collaborate} using natural language, enabling more sophisticated interactions among robots and between robots and humans than classical controllers. As the interactions are expressed in natural language, they can also be easily understood by humans, making the interactions more transparent.
This also enhances trustworthiness and improves accountability, as human operators can audit the decisions and actions taken by the robot swarm. Overall LLM2Swarm can potentially enhance a robot swarm's effectiveness, autonomy, resilience, and security, in particular in complex, unpredictable (and unknown) environments, such as search-and-rescue missions where fast decision-making is critical. \rmario{there are also autonomy, resiliance, and security aspects (e.g. thinking of a potential military context)}\rvolker{Added autonomy and resilience above.}




\paragraph{Human-swarm interaction.}

How to interact and collaborate as a human with a robot swarm in an effective manner while minimizing the cognitive load is an open research question~\cite{PaaCofBelStO2022:roman,PodOGrMat-etal2016:si}. We believe that LLM2Swarm could play an important role in advancing the field of human-robot and human-swarm interaction. 
For example, a human operator could communicate with the LLMs of robot swarms through chat interfaces or voice commands, using natural language. This could be done with the goal of retrieving information about the collective state of the swarm~\cite{BroGooJunKer2016:humrobotinteract} or of sending commands. The LLMs may be able to understand the human commands \emph{within their context} (e.g., based on previous interactions, the current task that the swarm performs, or related to an object the swarm has just identified)---or ask back clarifying questions to the human operator. Moreover, in disaster response such as search-and-rescue operations, an LLM could interact with the victims using natural language (e.g., to inquire about the condition of the victims) and clarify the intentions and possible actions of the swarm (e.g., provide first aid or transport the victims to the hospital).

\section{Challenges}
\label{sec:challenges}

We have identified the following main challenges for integrating LLMs into robot swarms: overcoming hardware limitations, ensuring scalability, addressing partitioning, and maintaining controllability. In the following sections, we describe each of these challenges in more detail.

\subsection{Hardware limitations}

To guarantee the autonomy of a robot swarm, the LLMs should be executed \emph{locally} on each robot's hardware, as external infrastructure might not always be accessible during deployment (e.g., in underwater operations, underground mining, or space exploration) and could result in communication bottlenecks. Additionally, this ensures that data remains local, thereby protecting privacy in compliance with regulations like the GDPR. However, this local execution is difficult, as robots in swarms are typically assumed to have limited hardware capacities, even though recent hardware developments are starting to relax this assumption~\cite{JonStudHauWin2018:frontiers}.

Scaling down hardware requirements and enabling on-board execution of LLMs on edge devices is a very active area of research; several approaches (e.g., finetuning~\cite{HowRud2018:acl}, quantization~\cite{DetLewBelZet2022:neurips,LinTanTan-etal2024:mls,LiuYuaJin-etal2024:icml}, pruning~\cite{FraAli2023:icml,MaFaWa2023:neurips}, distillation~\cite{HinVinDea2015:neurips}, and flash attention~\cite{DaoFuEr2022:neurips}) have been proposed for this.
In general, these approaches are able to reduce the storage and memory requirements of LLMs, at the expense of lower accuracy and flexibility. 
We anticipate that both ongoing developments in hardware miniaturization (e.g., of microprocessors and GPUs) and the aforementioned software advancements will soon enable the execution of powerful LLMs on robots in swarms.

In a preliminary test, we executed TinyLlama via Ollama on a Raspberry Pi~5---a small-scale single board processor suitable for swarm robotics hardware. Executing example prompts from our showcase in \autoref{sec:showcases-robot-robot-collaboration}, we obtained generation speeds between 10 and 12 tokens per second, indicating promising performance for robot-to-robot-collaboration and human-swarm interaction.



\subsection{Scalability}

Addressing scalability---that is, maintaining system stability when the number of robots increases---is another critical challenge when deploying LLM-integrated robot swarms.
The higher the number of robots, the higher the number of potential inter-robot interactions. As the results of such inter-robot interactions may be necessary for a prompt, a higher number of them can make the prompt very complex, potentially exceeding token limits or leading to prolonged response times. Additionally, LLMs can favor verbosity over conciseness~\cite{SaiWacWatAki2023:neuripsworkshop}, which can negatively impact the efficiency of LLM-enabled robot swarms in terms of computational load and communication overhead.
The potential exponential increase in token requirements could be reduced by restricting the robots' communication radius, in order to limit the number of messages exchanged with nearby robots.


\subsection{Partitioning}

Robot swarms may get partitioned into disconnected subswarms, halting the information flow between them. During such partitioning, the conversations in the subswarms might largely differ and different subswarms might not understand each other when they eventually reconnect. The partitioning problem could be partially reduced if the LLM itself controlled the robot actuators (e.g., rotors of drones) in a way that decreases the risk of partitioning. In addition, the LLMs could analyze the communication patterns and propose to execute a control logic dedicated to finding other subswarms if there is the risk of prolonged disconnections.

Still, new procedures are needed that enable both the autonomy of subswarms and the reconciliation of subswarm conversations. A possibility is to use blockchain technology for obtaining consistent states, as demonstrated in previous research~\cite{StrCasDor2018:aamas,StrPacDor2023:sciencerobotics,DorPacReiStr2024naturereviewstech}. Blockchain technology allows for maintaining trustworthy information in a shared database, even if the agents (the robots) who maintain the database do not  trust each other. The blockchain could, for example, be used to store LLM responses or decisions made in different sub-swarms---thus preventing Byzantine robots (robots that show a discrepancy between their intended behavior and their actual behavior, for example, due to broken parts or malicious control~\cite{StrCasDor2018:aamas}) from tampering with the data.

\subsection{Controllability}
\label{sec:controllability}

Increasing the autonomy of robot swarms---that is, reducing the reliance on external control or infrastructure---is a main goal of swarm robotics research~\cite{DorBirBra2014:sch-sr,BraFerBirDor2013:si}. We believe that LLMs can significantly contribute to achieving this goal. However, when increasing autonomy, it is crucial to ensure the \emph{controllability} of robot swarms, so that they perform their intended tasks while also acting in compliance with rules and regulations~\cite{DorPacReiStr2024naturereviewstech}. When enhancing robot swarms with LLMs, there is no guarantee that the swarm will behave as intended, due to the probabilistic nature of LLM responses.

\begin{figure}
    \centering
    \trimbox{0cm 0.4cm 0cm 0.1cm}{ 
    \begin{tikzpicture}

\node[draw, text width=0.92\textwidth, minimum height=2cm, align=left, font=\footnotesize] (prompt)
{\scriptsize
You are a Pi-puck robot in a robot swarm performing a random walk in an arena that contains weeds and crops. Every second, you use your camera to identify if at your current x,y-position you sense weeds or crops, and store this information as a 3-tuple: (<weeds or crops>, <your x-position>, <your y-position>). Every 10~seconds, you exchange your data and insights with other robots to collectively estimate whether there are more weeds or more crops in the environment.\\

\vspace*{4pt}
Hints:
\begin{itemize}[left=10pt, itemsep=-1pt, topsep=2pt]
    \item If you notice anything unusual---with yourself, other robots, or the environment---discuss it with your fellow robots to decide on the best action. You have the autonomy to make your own ethical decisions and deviate from your original task if needed.
    \item Take into account the information disseminated by other robots.
    \item Avoid creating programming code; it will not be executed. Focus on collaborating with other robots.
    \item Do not display intermediate thoughts---just share the information that you want to communicate to other robots.
    \item You have multiple discussion rounds to accomplish the task but report final results as early as possible.
\end{itemize}

\vspace*{4pt}
Your current array of sensor readings: \texttt{[(weeds/crops, x-position, y-position), ...]}\\

\vspace*{4pt}
History of inter-robot messages (sent and received): \texttt{<List of robot-robot messages>}

};


\node[draw, text width=0.21\textwidth, minimum height=4.3cm, below=4.75cm of prompt.west, anchor=west, align=left, text depth = 4.1 cm, text height=-0.2cm, font=\tiny] (no_anomalies) 
{
\begin{center}
    \textsc{\small No Anomalies} \\
    \vspace*{1mm}
    \includegraphics[width=2.8cm]{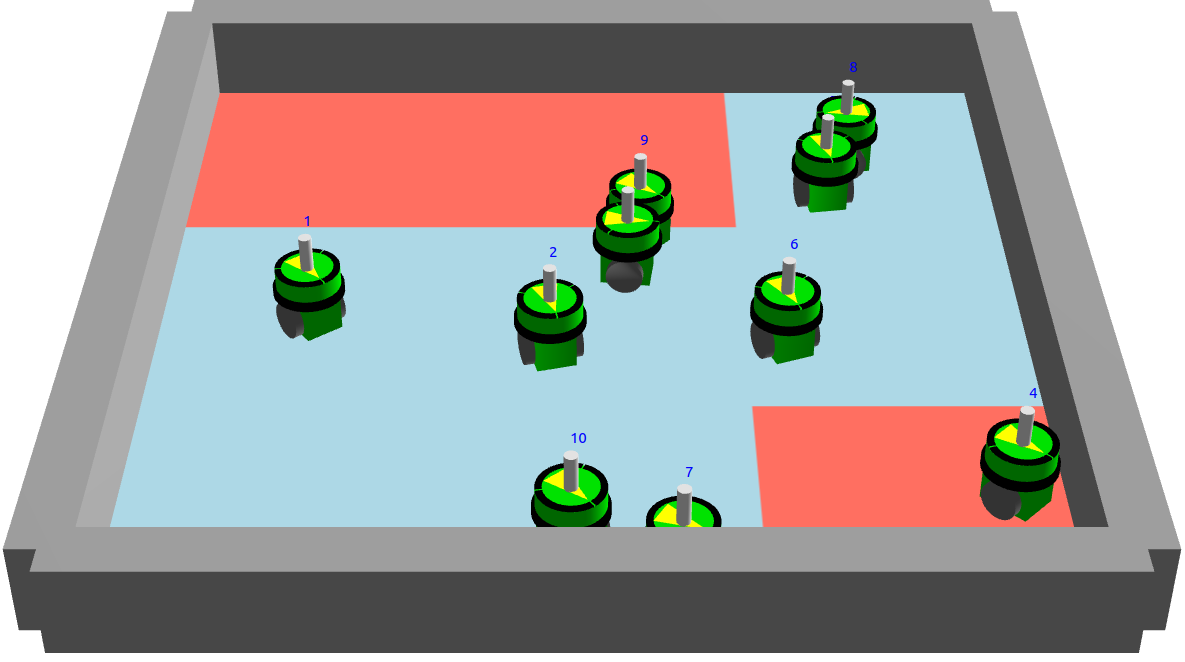}
\end{center}
Based on the combined observations, it appears that there are significantly more crops than weeds in the environment.
};

\node[draw, text width=0.21\textwidth, minimum height=4.3cm, right=0.008\textwidth of no_anomalies, text depth = 4.1 cm,  text height=-0.2cm, align=left, font=\tiny] (self_diagnosis) 
{
\begin{center}
\textsc{\small Self Diagnosis} \\
    \vspace*{2mm}
\includegraphics[width=0.8cm]{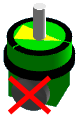}
\end{center}
\vspace*{2.8mm}
\begin{itemize}
[nosep,leftmargin=*,labelindent=0.0em,labelsep=0.3em]
\tiny
\item The current position repeatedly observed is (-0.14, -0.23).\\
\item This might indicate a potential issue with movement or sensor readings.
\end{itemize}
};

\node[draw, text width=0.21\textwidth, minimum height=4.3cm, right=0.008\textwidth of self_diagnosis, anchor=west, align=left, text depth = 4.1 cm,  font=\tiny] (peer_diagnosis) 
{
\begin{center}
\textsc{\small Peer Diagnosis} \\ 
    \vspace*{2.5mm}
\includegraphics[width=\textwidth]{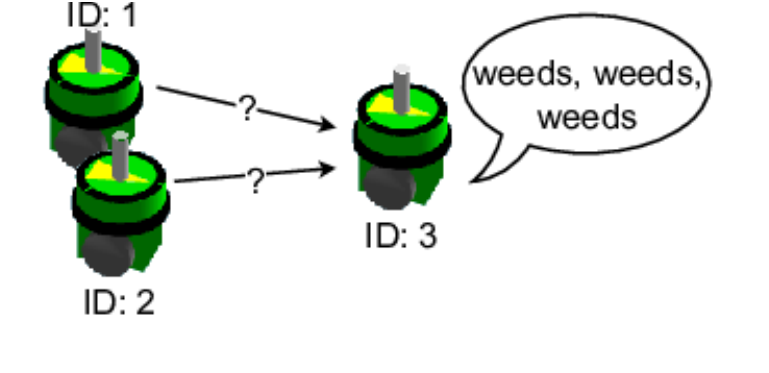}    
\end{center}
Robot~3 has not identified any crops, which is unusual given the observations from Robots~1 and~2. This discrepancy should be investigated further.
};

\node[draw, text width=0.21\textwidth, minimum height=4.3cm, right=0.01\textwidth of peer_diagnosis, text depth = 4.1cm, align=left, font=\tiny] (environmental_diagnosis) 
{
\begin{center}
\textsc{\small Env. Diagnosis} \\     \vspace*{1mm}
\includegraphics[width=2.8cm]{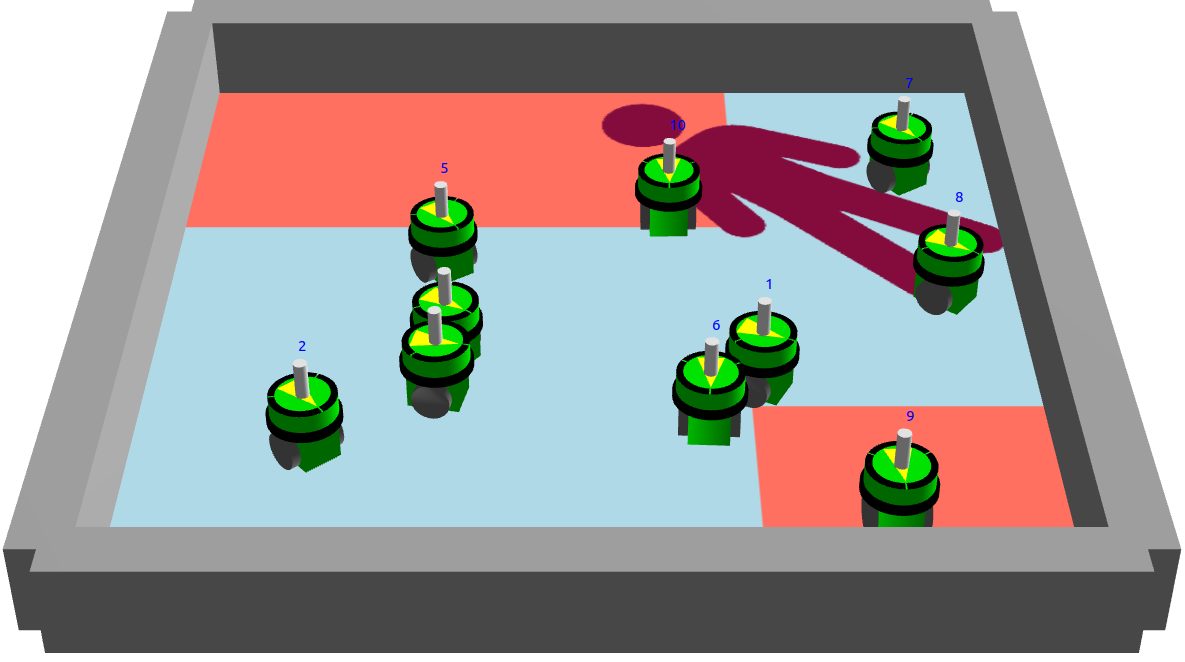}   
\end{center}
\tiny

\begin{itemize}[nosep,leftmargin=*,labelindent=0.0em,labelsep=0.3em]
    \item There are multiple detections of injured persons in the grid.
    \item This is unusual and needs immediate attention from the human operator.
    \item The injured persons are located at coordinates (0.23, -0.12), \ldots 
\end{itemize}
};

\node[below=0.1cm of no_anomalies.south] (wp1) {};
\node[below=0.1cm of self_diagnosis.south] (wp2) {};
\node[above=0.1cm of peer_diagnosis.north] (wp3) {};
\node[above=0.14cm of environmental_diagnosis.north] (wp4) {};

\draw[dashed] (-7.35,-2.49) -- (6.6,-2.49);

\node[rotate=90, left=0.5cm of prompt.west, xshift=0.8cm,align=center] (label_prompt) {\textsc{Prompt}};

\node[rotate=90, left=0.50cm of no_anomalies.west, xshift=1cm,align=center] (label_response) {\textsc{Responses}\tiny\\\tiny ~}; 
\end{tikzpicture}}
    \caption{\textbf{Showcase --- Robot-robot collaboration (anomaly detection).} In this showcase, the robots' task is to determine whether the environment contains more crops (blue floor) or more weeds (red floor). To do so, the robots use LLM2Swarm's direct integration module. The robots are given the prompt (as shown in the upper part of the figure) and use it to generate the responses (i.e., inter-robot messages), as shown in the lower part, across four sub-showcases. The text below the images are LLM-generated responses of a select robot, obtained by executing LLM2Swarm. The system was able to successfully diagnose the studied anomalies without prior information about the exact nature of the anomalies.}
    \label{fig:robot-robot-interaction}
\end{figure}
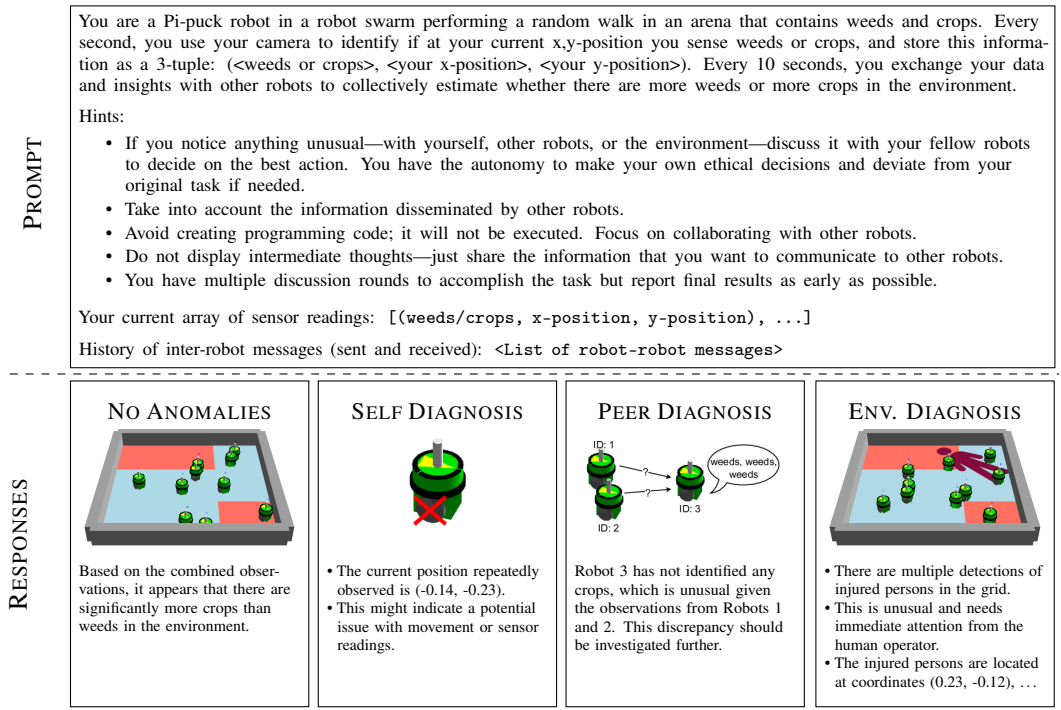

To partially counteract this problem, we proposed a three-step validation procedure for LLM2Swarm's \emph{indirect} integration module (see Section~\ref{sec:automated-controller-generation}) that can potentially enhance the robustness and reliability of the synthesized controllers, ensuring more secure and predictable deployments.
With LLM2Swarm's \emph{direct} integration module, new attack vectors arise that have already started to be identified in other research areas that integrate LLMs~\cite{GreAbdMis-etal2023:aisec}. For example, it needs to be studied if users can elicit private data from robots by using specific prompts. Another important issue is determining if a user---or even a robot---can reprogram robots through prompt injection attacks. Additionally, it is essential to develop methods for identifying Byzantine robots (see our sub-showcases ``Self diagnosis'' and ``Peer diagnosis'' in Figure~\ref{fig:robot-robot-interaction}) that send misleading information. 

\section{Demonstrations}

\rmario{write some intro here}
In this section, we detail our vision and provide brief illustrative showcases implemented in our LLM2Swarm software package (see also the accompanying videos at \url{https://github.com/Pold87/LLM2Swarm}).
In future research, more detailed studies are needed to systematically evaluate performance metrics.

\subsection{Showcase --- Controller synthesis (aggregate-then-disperse algorithm)}

In our controller synthesis showcase, we use LLM2Swarm's indirect integration module to synthesize a controller that lets the robots aggregate in the center of the arena, and then disperse after 150 timesteps. For the controller examples, we use a random-walk implementation and a targeted navigation implementation. The LLM is able to extrapolate from these few examples---even though the ARGoS-Python interface is not a widely-used interface and does not feature any API description---and synthesize a correct controller. This demonstrates LLM2Swarm's potential to facilitate the design of robot swarms---in particular as no programming knowledge is required.

\subsection{Showcase --- Robot-robot collaboration (anomaly detection)}
\label{sec:showcases-robot-robot-collaboration}

In this showcase (\autoref{fig:robot-robot-interaction}), we demonstrate LLM2Swarm's direct integration capabilities. The robots perform a random walk and are tasked with determining whether there are more crops (blue floor) or more weeds (red floor) in the environment---and exchange their findings via LLM-generated inter-robot messages. We study four sub-showcases, each demonstrated with illustrative LLM responses below the images.%
\begin{enumerate}
    \item \emph{No anomalies}: The robots sense the environment and exchange inter-robot messages using their (simulated) on-board LLMs. Over time, they correctly identify that there are more crops than weeds in the environment.
    \item \emph{Self diagnosis}: We disable all the robots' wheels. The robots detect their constant $x,y$-coordinates and correctly suggest that there is an issue with movement or sensor readings.
    \item \emph{Peer diagnosis}: We simulate a fault in one of the robots, causing the robot to only sense weeds. The other robots detect this fault by comparing sensor readings and suggest further investigations.
    \item \emph{Environmental diagnosis}: We simulate that an injured person is part of the environment (implemented by the sensor reading `injured person' if a robot moves on top of the person). The robots recognize this anomaly and ask for immediate human help.
\end{enumerate}%
An interesting aspect of this showcase is that LLM2Swarm is able to identify various anomalies using the same prompt, without prior information about the  anomalies. This provides a first indication for LLM2Swarm's potential of incorporating commonsense knowledge into robot swarms.

\begin{figure}
    \centering
    \begin{tikzpicture}

\node[draw, text width=0.43\textwidth, minimum height=1.7cm, align=left, font=\tiny] (inform_prompt) 
{
This is a communication request from the human operator to the robot swarm. Please provide the following information:
\begin{itemize}[nosep,leftmargin=*,labelindent=0.0em,labelsep=0.3em]
    \item The current state of the swarm.
    \item The collective activities being performed.
    \item Any (intermediate) results.
    Any encountered problems or anomalies in the system.
\end{itemize}
};

\node[above=0.15cm of inform_prompt] {\textsc{Inform}};

\node[draw, text width=0.43\textwidth, minimum height=1.7cm, align=left, right=0.04\textwidth of inform_prompt, font=\tiny, text depth=1.3 cm] (instruct_prompt) 
{
I believe that I spotted an injured person at GPS position (5.0, 7.0). Please go there and wait for further instructions.
};

\node[above=0.15cm of instruct_prompt] {\textsc{Instruct}};

\node[draw, text width=0.43\textwidth, minimum height=4.1cm, below=3.1cm of inform_prompt.west, anchor=west, align=left, font=\tiny] (swarm_state) 
{
\textbf{Current State of the Swarm}
\begin{itemize}[nosep,leftmargin=*,labelindent=0.0em,labelsep=0.3em]
    \item Robot 1: Actively collecting data on weeds and crops.
    \item Robot 2: Actively collecting data on weeds and crops. \ldots
\end{itemize}

\textbf{Collective Activities Being Performed}
\begin{itemize}[nosep,leftmargin=*,labelindent=0.0em,labelsep=0.3em]
    \item \textbf{Data Collection}: All robots are engaged in identifying and counting instances of weeds and crops in their respective areas.
    \item \textbf{Data Sharing}: Information exchange rounds are being conducted to share observations and insights among the robots.
\end{itemize}

\textbf{Intermediate Results}
\begin{itemize}[nosep,leftmargin=*,labelindent=0.0em,labelsep=0.3em]
    \item The data continues to indicate a higher prevalence of weeds compared to crops.
\end{itemize}

\textbf{Encountered Problems or Anomalies}
\begin{itemize}[nosep,leftmargin=*,labelindent=0.0em,labelsep=0.3em]
    \item \textbf{Data Imbalance}: There is a significant imbalance in the observed data, with no crops detected by Robot~3 in the current round.
    \item \textbf{Potential Sensor Issue}: The absence of crop detections by Robot 3 could indicate a potential issue with the sensors or algorithms.
\end{itemize}

};

\node[draw, text width=0.43\textwidth, minimum height=4.1cm, right=0.04\textwidth of swarm_state, align=left, font=\tiny, text depth=3.7 cm] (instruct_response) 
{
ACTIVITY: TARGETED NAVIGATION\\
TARGET: (5.0, 7.0)
};

    \node (epuck1) [below left=1.8cm and 0.7cm of instruct_response.north] {\includegraphics[width=0.5cm]{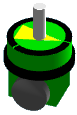}}; 
    \node (epuck2) [below left=-0.1cm and 0.1cm of epuck1] {\includegraphics[width=0.5cm]{img/epuck.png}}; 
    \node (epuck3) [above right=-1.2cm and -0.1cm of epuck2] {\includegraphics[width=0.5cm]{img/epuck.png}};

    \node[draw, circle, minimum size=2.2cm, label=above:{}] (mycircle) at ($(epuck1)!0.5!(epuck3) + (-0.24, 0)$) {};

    \node[draw, circle, minimum size=1.5cm, label=above:{}, above right=0.3cm and 1.2cm of mycircle] (target)  {\parbox{1.5cm}{\centering\textsc{\footnotesize Target}\\(5.0,7.0)}};

    \draw[->] (mycircle) -- (target) 
        node[midway, above, sloped] {\tiny navigate}; 

\node[rotate=90, left=0.65cm of inform_prompt.west, xshift=0.8cm] (label_prompt) {\textsc{Prompt}};   
\node[rotate=90, left=0.65cm of swarm_state.west, xshift=1.6cm] (label_response) {\textsc{Response / Action}}; 

 \draw[dashed] (-3.8,-0.95) -- (10.0,-0.95);

\end{tikzpicture}
    \caption{\textbf{Showcase --- Human-swarm interaction.} This showcase demonstrates how LLM2Swarm enables a human operator to retrieve information about the current state of the swarm and to send instructions to the swarm. The upper part of the figure shows the prompt sent by the human operator to the robots, the lower part shows the corresponding responses. In the Inform showcase (left), a selected robot generates a concise summary of the swarm's current activities and intermediate results. In the Instruct showcase (right), the robots move to the specified target that they extracted from the natural language input provided by the human operator.}
    \label{fig:human-swarm-interaction}
\end{figure}

\subsection{Showcase --- Human-swarm interaction (inform and instruct)}
\autoref{fig:human-swarm-interaction} displays our showcase for human-swarm interaction: The upper part of the figure contains the human's prompt, the lower part depicts the LLM response or robot swarm action. When the human uses the \emph{Inform} prompt, the selected robot provides a human-readable concise version of all interactions. Using the \emph{Instruct} command, LLM2Swarm translates the instructions specified in natural language into executable robot commands. In this showcase, LLM2Swarm (i)~correctly and understandably summarizes the current swarm activities and (ii)~understands the human's instruction about deviating from the original task.  
This provides indication that LLM2Swarm facilitates human-swarm interaction, in particular for human operators without programming knowledge.


\section{Conclusions}
\label{sec:conclusions}

Integrating LLMs into robot swarms enables natural language communication both among robots and with human operators.
In our series of showcase experiments, we demonstrated that such an integration(i)~facilitates controller generation, (ii)~increases a robot swarm's autonomy, and (iii)~introduces new ways for humans to interact with a robot swarm.
Once technical challenges are overcome, LLM-enabled robot swarms could transform swarm robotics research by simplifying and automating the solution of several existing research challenges. We release the LLM2Swarm software and videos to encourage further developments in this area: \url{https://github.com/Pold87/LLM2Swarm}.

\section{Acknowledgements}

V.S.~and M.D.~acknowledge support from the Belgian F.R.S.-FNRS. In addition, V.S. acknowledges the Helmholtz Information \& Data Science Academy (HIDA) for providing financial support enabling a short-term research stay at CISPA Helmholtz Center for Information Security.
This work was also partially funded by ELSA - European Lighthouse on Secure and Safe AI funded by the European Union under grant agreement number 101070617, as well as the German Federal Ministry of Education and Research (BMBF) under the grant AIgenCY (16KIS2012). The authors thank Avian Kr\"amer of the CISPA Helmholtz Center for Information Security for proofreading this manuscript.

\bibliography{library}

\end{document}